# Reinforcement Learning with Partially Known World Dynamics

**Christian R. Shelton**
Computer Science Department
Stanford University
cshelton@cs.stanford.edu

## Abstract

Reinforcement learning would enjoy better success on real-world problems if domain knowledge could be imparted to the algorithm by the modelers. Most problems have both hidden state and unknown dynamics. Partially observable Markov decision processes (POMDPs) allow for the modeling of both. Unfortunately, they do not provide a natural framework in which to specify knowledge about the domain dynamics. The designer must either admit to knowing nothing about the dynamics or completely specify the dynamics (thereby turning it into a planning problem). We propose a new framework called a partially known Markov decision process (PKMDP) which allows the designer to specify known dynamics while still leaving portions of the environment's dynamics unknown. The model represents not only the environment dynamics but also the agent's knowledge of the dynamics. We present a reinforcement learning algorithm for this model based on importance sampling. The algorithm incorporates planning based on the known dynamics and learning about the unknown dynamics. Our results clearly demonstrate the ability to add domain knowledge and the resulting benefits for learning.

## 1 Introduction

Reinforcement learning extends the tantalizing promise of being able to automatically learn a method of acting in a dynamic world that will maximize the long-term value for any reward function and any unknown dynamics using only experience in the world. Unfortunately, this problem specification is often too general. Without any additional information, the task of finding an optimal policy (or even a good policy) is daunting. Given that the environment in which the agent has to act might have arbitrary dynamics, the agent must entertain the possibility that the underlying state-space of the environment has an exponential (in the length of the time sequence) number of states.

Complete observability is the most common assumption made to simplify the problem. This greatly increases the tractability of learning as each time step can be considered on its own and the context of the trajectory from which it came can be ignored. The resulting decomposition leads to great gains in data efficiency. While there still are many open problems for the complete observability case, in this paper we would like to allow for unobserved state in the underlying state dynamics. We feel that most practical problems for reinforcement learning have hidden state that is essential to the problem formulation.

Two methods for coping with hidden state have arisen. The first is gradient ascent methods (Williams, 1992). Instead of explicitly modeling the exponential number of hidden states, gradient ascent algorithms fix a parameterized policy class and attempt to optimize the expected return by local hill-climbing within the class. By searching for the solution in a restricted set and settling for a suboptimal solution, the problem becomes simpler.

The second method is to assume the dynamics of the system are known. The agent's job is then no longer one of learning, but of planning. Even so, the problem of finding the optimal policy is PSPACE-hard (Littman, 1996). However, this does eliminate the need to gather data about a partially unobserved process. Approximation and local search algorithms can often provide good solutions in this context (Rodríguez et al., 2000).

We are unwilling to require that the dynamics of the system be completely known. However, it is often the case that portions of the environment are known to the system designers. For example, the dynamics of a robot (excepting the influences from a few unknown world forces) may be well understood and modeled: we might know that if the robot issues the command to move forward either it will move forward successfully in which case the distribution over final position is known, or it will run into an obstacle in which case it will end up next to the obstacle. There is still an unknown component to the model corresponding to whether an obstacle will impede the robot's motion. However, we should not be forced to conclude that we must allow arbitrary unknown dynamics or that we must come up with a model to explain when and how obstacles will be encountered. A reinforcement learning algorithm should be able to learn about the dynamics of the obstacles while knowing (and not learning) about the dynamics of its locomotion system. We desire the ability to incorporate our limited knowledge about the environment while still allowing the algorithm to learn about the dynamics of environment that are outside of our knowledge.

In this paper we present a framework that allows the incorporation of some forms of such information and an algorithm for exploiting it in a reinforcement learning setting. It is an extension of the POMDP model allowing for both planning and learning by making the agent's knowledge of the dynamics explicit.

There are a number of currently known techniques to add domain knowledge to reinforcement learning. Reward shaping (Ng et al., 1999) can be used to change the return function to yield easier value functions without changing the optimal policy.. Action restriction can been used to limit the search space based on domain knowledge, either by external advice (Maclin & Shavlik, 1996), hierarchies (Parr, 1998; Precup, 2000; Dietterich, 2000), or Lyapunov functions (Perkins & Barto, 2001). It is also common with any technique involving function approximation to construct the function approximator to take advantage of knowledge about the invariants in the value function. Nikovski and Nourbakhsh (2000) show some methods for learning in partially-observable domains that allow the incorporation of known parameters. However, most assume the state-space of the unobserved variables is known. Shelton (2001) is a specific case of the model in this paper for the case of memory bit dynamics. We extend those ideas to general knowledge about world dynamics and allow the reward function to depend both on modeled and unmodeled portions of the environment.

This paper presents a different approach. We allow the domain knowledge to take the form of a probability model for portion of the world that is conditional on the behavior of the rest of the environment. We divide the world into the known part (for which we have a probabilistic model) and the unknown part (for which we do not). This formulation is inherently a partially-observable model and naturally takes into account the differences among the observed, modeled, and unmodeled state-spaces.

## 2 PKMDP Formulation

Figure 1 shows the partially known Markov decision process (PKMDP) as a graphical model. We divide the hidden state of the world into two parts. The first part, denoted by the variable $s$, we call the unknown state. Its dynamics are not known to the learning algorithm. The second part, denoted by the variable $x$, we call the known state. Its dynamics are given to the algorithm. However, the label "known" refers solely to its dynamics; it is not directly observed by the agent. $y$ represents the effect of the unknown state on the known state and $z$ represents the effect of the known state on the unknown state. Together $y$ and $z$ can be viewed as the "interface" between the known dynamics and the unknown part of the environment. $y$ is the information from the unknown environment necessary to update the known portion, and $z$ is the information from the known portion necessary to update the unknown environment. $o$ is observation of the agent and $a$ is the action performed by the agent. Not drawn in the figure are the reward values. With each value for $s$ and, separately, for each value of $x$ is associated a real-valued reward.

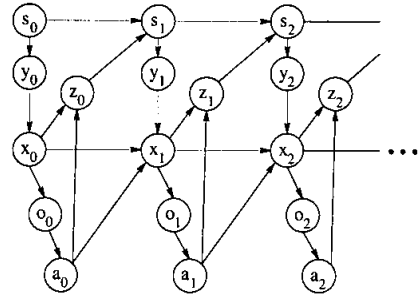

Figure 1: Graphical representation of the PKMDP model.

For example, in our robot environment from the previous section, $s$ would represent the obstacles in the environment. Their position, velocity, number, and dynamics are all unknown in both quantity and representation. The variable $x$ would represent the position, velocity, and other associated state of the robot, all of which are known and well-modeled. The variable $y$ would represent the effect of the obstacles on the robot. In a simple case, this might be whether the current robot action was interrupted by an obstacle and, if so, where. The variable $z$ would represent the effect of the robot on the obstacles. It might just be the position of the robot, if we are willing to assume the other state of the robot (internal sensor readings and velocities) do not change the obstacles' behaviors. As a normal POMDP, $o$ represents the sensor readings and $a$ is the action chosen by the robot.

Before specifying the model more precisely, we should note a few features of this model. If the variables $s$, $y$, and $z$ are degenerate (*i.e.* they have only one possible value), this model is the same as a POMDP model used for planning ($x$ is the hidden state). Similarly, if the variable $x$ is a copy of $y$ (*i.e.* the state space of $x$ is the same as that of $y$ and the conditional probability function associated with $x$ is an identity mapping and depends only on $y$) and the variable $o$ is similarly a copy of $x$, this model is the same as a POMDP model used for learning ($s$ is the hidden state).

A partially known Markov decision process (PKMDP) is a 8-tuple, $\langle S, Y, Z, X, O, A, P, R \rangle$. $S$, $Y$, $Z$, $X$, $O$, and $A$ are the sets of possible values for the variables $s$, $y$, $z$, $x$, and $o$ respectively. $P$ is a set of conditional probability distributions ($p_s$, $p_y$, $p_z$, $p_x$, and $p_o$) for these same variables given their parents in the graph of figure 1 and two starting distributions ($p_{s0}$ and $p_{x0}$) over $s_0$ and $x_0$ (given $y_0$). $R$ is a pair of deterministic real-valued reward functions, $r_s$ and $r_x$: the rewards associated with states $s$ and $x$ respectively.

With a policy[1], a distribution over $a$ given $o$ which we will denote $\pi$, this forms a complete distribution over state sequences of fixed length. The generative model is to draw the variables $s$, $y$, $x$, $o$, $a$, and $z$ in order for a particular time step according to the conditional probability distributions for each variable. This is a two time-slice Bayesian network (2-TBN) over six variables.

### 2.1 Episodic Reinforcement Learning

For reinforcement learning, we assume that the agent knows the PKMDP with the exception of $S$, $p_y$, $p_s$, $p_{s0}$, and $r_s$. For the episodic case which we will be concentrating on in this paper, the agent is repeatedly restarted at the beginning of the PKMDP (*i.e.* time is reset to 0). For each restart, the agent is allowed to select a different policy (using the data gathered from previous restarts). The episode is run using this policy until either a terminate state is reached or a fixed number of time steps have passed. At each time step, the agent preserves the observation, takes an action, and receives the reward associated with the state $s$ and the reward associated with the state $x$. In contrast to standard reinforcement learning models, after the entire trial is complete the agent is also shown the values of the variables $y$ and $z$ for the duration of the trial. The goal of the agent is to select policies based on previous trials to maximize the sum of the rewards given by the unknown state $s$ and the rewards given by the known state $x$:

$$R(\pi) = E_\pi \left[ \sum_t r_s(s_t) + r_x(x_t) \right]$$

where $\pi$ is the policy chosen. The expectation is with respect to the distribution over state trajectories implied by the combination of the PKMDP model and the policy $\pi$.

Without observing $y$ and $z$, the task becomes significantly more difficult; the difference between unknown and known state is blurred. While it may be possible to formulate useful algorithms when they are hidden, for this paper we will concentrate on the case where these variables are observed at the end of an episode. Section 4 demonstrates how this restriction plays out in a couple of examples.

## 3 Algorithm

We propose a greedy reinforcement learning algorithm. The goal is to find the optimal policy within a parameterized space of policies. Before each episode, the algorithm uses the data from the previous episodes to construct an estimate of the expected return for any new policy. Using local optimization, the algorithm maximizes this estimate with respect to the policy parameters. The episode is run using this maximized policy. The policy used and data collected from the episode are added to the total data of the algorithm and the process repeats. A more general discussion of such algorithms based on importance sampling can be found in Peshkin and Shelton (2002).

This section is divided into three parts. The first part outlines importance sampling in its general form. The second part shows how to construct an importance sampling estimate for a new policy. There are two important points in this construction. The first is that the estimator requires no knowledge of the unknown state space $S$, the sampled $s$ sequence, or the distributions $p_s$ and $p_{s0}$, all of which are not known to the agent. The second crucial point is that this estimate incorporates the knowledge that the agent does have of the dynamics of the known state $x$. Therefore, the agent is not using sampling to estimate known distributions.

We use conjugate gradient ascent to optimize the return estimate. The final part of this section shows how the quantities necessary for the optimization can be calculated in time that is linear in the length of the episode. This is based on simple dynamic programming in the style of hidden Markov model inference. These calculations embody the algorithm's planning within the known part of the model.

### 3.1 Importance Sampling

The expected return estimate must be constructed from data collected under policies that differ from the policy to be evaluated (and therefore have different distributions over state trajectories). Importance sampling (or likelihood ratios) are well-suited to this task.

Importance sampling allows the estimation of an expectation with respect to one distribution from samples from a different distribution. In particular if samples $x^1, x^2, \ldots, x^n$ were each drawn i.i.d. from the distribution $p$,[2] the importance sampling estimator

$$\hat{f} = \frac{1}{n} \sum_i f(x^i) \frac{q(x^i)}{p(x^i)}$$

---

[1] For the moment, we are only allowing reactive policies, or policies without memory. In section 4 we demonstrate how memory can be folded into the variable $x$.

[2] We use superscripts to denote samples to keep them separate from subscripts used to denote time.

is an unbiased estimate of $E_q[f(x)]$ if $p(x)$ and $q(x)$ are positive. The weighted importance sampling estimator

$$\hat{f} = \frac{\sum_i f(x^i)\frac{q(x^i)}{p(x^i)}}{\sum_i \frac{q(x^i)}{p(x^i)}}$$

has smaller variance at the expense of adding slight bias. In practice it performs much better.

If the data were drawn from a set of different distributions, we could still perform this estimate. In particular, if sample $x^i$ were drawn from distribution $p^i$, then the above estimates still work if we define

$$p(x) = \frac{1}{n}\sum_j p^j(x) \ .$$

For more information on importance sampling in general, we refer to Rubinstein (1981) and Hesterberg (1995). Peshkin and Mukherjee (2001), Precup et al. (2000; 2001), and Shelton (2001) study these estimators in the context of reinforcement learning.

### 3.2 Return Estimation

In a reinforcement learning setting, the distributions $p$ and $q$ from the previous section correspond to different distributions over trajectories induced by different policies. We require all policies to assign a positive probability to every action choice from every observation to insure the importance sampling estimator is well-defined. For the moment, we will assume that all of our previous data samples were obtained by the same policy $\pi'$ and that our goal is to estimate the expected return for a second policy $\pi$. Furthermore, we will use unweighted importance sampling. Both allow for simpler expressions. At the end of this section, we will extend the expression to weighted importance sampling with varying sampling policies.

We let $b_{i:j}$ stand for the sequence $b_i, b_{i+1}, \ldots, b_j$ for any variable $b$. We let $T$ be the length of the sequence and $B$ stand for the entire sequence, $b_{0:T}$.

We must make a key preliminary observation. In particular, the probability of an entire trajectory can be split into two factors:[3]

$$p(S, Y, Z, X, O, A|\pi)$$
$$= \prod_t \begin{bmatrix} p_s(s_t|s_{t-1}, z_{t-1})p_y(y_t|s_t)p_x(x_t|x_{t-1}, y_t, a_{t-1}) \\ p_o(o_t|x_t)p_a(a_t|o_t, \pi)p_z(z_t|x_t, a_t) \end{bmatrix}$$
$$= \prod_t [p_s(s_t|s_{t-1}, z_{t-1})p_y(y_t|s_t)]$$
$$\times \prod_t \begin{bmatrix} p_x(x_t|x_{t-1}, y_t, a_{t-1})p_o(o_t|x_t) \\ p_a(a_t|o_t, \pi)p_z(z_t|x_t, a_t) \end{bmatrix}$$
$$= U(S, Y, Z) \times K(Y, Z, X, O, A, \pi) \ .$$

---
[3]We are ignoring the special case of $t = 0$ which splits just like every other time slice.

The last equality is by definition. The factor $U(S, Y, Z)$ contains all of the distributions that are unknown to the agent while the factor $K(Y, Z, X, O, A, \pi)$ contains the agent's policy and the known distributions.

Note that the factor $K(Y, Z, X, O, A, \pi)$ is exactly the probability of the sequences $Z$, $X$, $O$, and $A$ given the sequence $Y$ and the distribution $\pi$ in graph of figure 1 where the $s$ nodes and all outgoing or incoming edges to these nodes have been removed. We call this mutilated graph the severed model. We abuse notation and let $K(Y, Z, \pi) = \sum_{X,O,A} K(Y, Z, X, O, A, \pi)$, which is the probability of $Z$ given $Y$ in the severed model. These are subtly, but importantly, different from the same probabilities in the original PKMDP model. The probability of the sequences $S, Z, Y$ in the original PKMDP model given a policy $\pi$ can be written as $U(S, Y, Z)K(Y, Z, \pi)$.

The expected return can be broken into two parts: the sum of the rewards from the unknown state $s$ and the sum of the rewards from the known state $x$. We will tackle each in turn. The expected return from the unknown state is

$$E_\pi[R_s(S)] = \sum_{S,Y,Z} R_s(S)p(S, Y, Z|\pi)$$

where we have let $R_s(S) = \sum_t r_s(s_t)$. Taking our samples of $\langle S^i, Y^i, Z^i \rangle$ triplets and using importance sampling, we arrive at the estimate

$$\hat{R}_S(\pi) = \frac{1}{n}\sum_i R_s(S^i)\frac{U(S^i, Y^i, Z^i)K(Y^i, Z^i, \pi)}{U(S^i, Y^i, Z^i)K(Y^i, Z^i, \pi')}$$
$$= \frac{1}{n}\sum_i R_s(S^i)\frac{K(Y^i, Z^i, \pi)}{K(Y^i, Z^i, \pi')} \ .$$

This is computable: the return from the unknown state sequence as well as the $Y$ and $Z$ sequences are observed and the expression for $K$ contains only known probability distributions.

The rewards from the known state are slightly more difficult to tease into a similar expression.

$$E_\pi[R_x(X)]$$
$$= \sum_{S,Y,Z,X,O,A} R_x(X)p(S, Y, Z|\pi)p(X, O, A|Y, Z, \pi)$$
$$= \sum_{S,Y,Z} \left[\sum_{X,O,A} R_x(X)p(X, O, A|Y, Z, \pi)\right] p(S, Y, Z|\pi)$$

Again, we can apply importance sampling to use our data to estimate this quantity. Whereas before we were estimating the expectation of $R_s(S)$ with respect to the distribution over $S$, $Y$, and $Z$, here we are estimating the expectation of the quantity inside the brackets. The result is

$$\hat{R}_X(\pi) = \frac{1}{n} \sum_i \left[ \sum_{X,O,A} R_x(X) p(X,O,A|Y^i, Z^i, \pi) \right.$$
$$\left. \times \frac{K(Y^i, Z^i, \pi)}{K(Y^i, Z^i, \pi')} \right] .$$

If we now note that $p(X, O, A|Y, Z, \pi)$ is the same in the PKMDP model as it is in the severed model and remember that $K(Y, Z, \pi)$ is the probability of $Z$ given $Y$ in the severed model, we can finish the derivation. We let $\bar{p}$ denote probabilities in the severed model.

$$\hat{R}_X(\pi) = \frac{1}{n} \sum_i \left[ \sum_{X,O,A} R_x(X) \bar{p}(X,O,A|Y^i, Z^i, \pi) \right.$$
$$\left. \times \frac{\bar{p}(Z^i|Y^i, \pi)}{K(Y^i, Z^i, \pi')} \right]$$
$$= \frac{1}{n} \sum_i \frac{\sum_{X,O,A} R_x(X) \bar{p}(Z^i, X, O, A|Y^i, \pi)}{K(Y^i, Z^i, \pi')} .$$

We let $V(Y, Z, \pi)$ denote the quantity in the numerator,

$$V(Y, Z, \pi) = \sum_{X,O,A} R_x(X) \bar{p}(Z^i, X, O, A|Y^i, \pi) .$$

which is the portion of the expectation of the known return corresponding to when the sequence $Z$ is generated. The final expression for the estimator of both the known and unknown returns is

$$\hat{R}(\pi) = \frac{1}{n} \sum_i \frac{R_s(S^i) K(Y^i, Z^i, \pi) + V(Y^i, Z^i, \pi)}{K(Y^i, Z^i, \pi')} .$$

Translating this to the full case where trial $i$ used policy $\pi^i$ and weighted importance sampling, the estimate becomes

$$\hat{R}(\pi) = \frac{\sum_i \frac{R_s(S^i) K(Y^i, Z^i, \pi) + V(Y^i, Z^i, \pi)}{\sum_j K(Y^i, Z^i, \pi^j)}}{\sum_i \frac{K(Y^i, Z^i, \pi)}{\sum_j K(Y^i, Z^i, \pi^j)}} .$$

### 3.3 Dynamic Programming

To optimize $\hat{R}(\pi)$ with respect to the parameters of $\pi$, we need to be able to calculate $K(Y, Z, \pi)$, $V(Y, Z, \pi)$, and their derivatives with respect to $\pi$ efficiently. Fortunately, this is almost exactly like calculating the probability of an input-output hidden Markov model where $x$ is the hidden state, $y$ is the input, and $z$ is the observation (or output). We quickly review the needed formulas, but refer to Rabiner (1989) and Bengio (1999) for more complete discussions.

Both $K$ and $V$ are defined in terms of the probabilities of the severed model. For this section, all discussion will be for the severed model and we will drop the dependence on $\pi$ from the notation. We first define the matrix $T(y, z)$ to be the transition matrix for the known state, $x$, conditioned on the value $y$ and multiplied by the probability of producing the value $z$ during the transition. More formally, the element of $T(y, z)$ indexed by $x$ and $x'$ is

$$T_{x,x'}(y, z) = \sum_{o,a} p_o(o|x) p_a(a|o) p_z(z|x, a) p_x(x'|y, x, a) .$$

We can define the usual forward-backward equations.

$$\alpha_t(x) = p(x_t, z_{1:t-1}|Y)$$
$$\beta_t(x) = p(z_{t:T}|x_t, Y)$$

have the recursions

$$\alpha_0(x) = p_{x0}(x|z_0)$$
$$\alpha_{t+1}(x') = \sum_x T_{x,x'}(y_{t+1}, z_t) \alpha_t(x)$$
$$\beta_T(x) = \sum_{a,o} p_o(o|x) p_a(a|o) p_z(z_T|x, a) .$$
$$\beta_{t-1}(x) = \sum_{x'} T_{x,x'}(y_t, z_{t-1}) \beta_t(x')$$

This allows us to calculate in $O(T)$ time the probability of the $Z$ sequence given the $Y$ sequence, $K(Y, Z, \pi)$, and the expectation of the return scaled by the probability of the $Z$ sequence given the $Y$ sequence, $V(Y, Z, \pi)$.

$$K(Y, Z, \pi) = \sum_x \alpha_t(x) \beta_t(x) \qquad \forall t \in \{0, 1, \ldots T\}$$
$$V(Y, Z, \pi) = \sum_t \sum_x r_x(x) \alpha_t(x) \beta_t(x)$$

The derivation of the derivatives useful for optimization is too long to include here. It is sufficient to calculate the derivative with respect to $p_a(a|o)$ for all $a$ and $o$; if the policy has parameters other than these, their derivatives can be calculated via the chain rule. The necessary preliminary definitions are

$$\frac{\partial T_{x,x'}(y, z)}{\partial p_a(a|o)} = p_o(o|x) p_z(z_t|x, a) p_x(x'|y_{t+1}, x, a)$$

$$\frac{\partial \alpha_{t+1}(x')}{\partial p_a(a|o)} = \sum_x \left[ \begin{array}{l} T_{x,x'}(y_{t+1}, z_t) \frac{\partial \alpha_t(x)}{\partial p_a(a|o)} \\ + \frac{\partial T_{x,x'}(y_{t+1}, z_t)}{\partial p_a(a|o)} \alpha_t(x) \end{array} \right]$$

$$\frac{\partial \beta_{t-1}(x)}{\partial p_a(a|o)} = \sum_{x'} \left[ \begin{array}{l} T_{x,x'}(y_t, z_{t-1}) \frac{\partial \beta_t(x')}{\partial p_a(a|o)} \\ + \frac{\partial T_{x,x'}(y_t, z_{t-1})}{\partial p_a(a|o)} \beta_t(x') \end{array} \right]$$

which allow

$$\frac{\partial K(Y, Z, \pi)}{\partial p_a(a|o)} = \sum_t \sum_{x,x'} \alpha_t(x) \frac{\partial T_{x,x'}(y_{t+1}, z_t)}{\partial p_a(a|o)} \beta_{t+1}(x')$$

$$\frac{\partial V(Y, Z, \pi)}{\partial p_a(a|o)} = \sum_x r_x(x) \sum_t \begin{array}{l} \alpha_t(x) \frac{\partial \beta_t(x)}{\partial p_a(a|o)} \\ + \beta_t(x) \frac{\partial \alpha_t(x)}{\partial p_a(a|o)} \end{array} .$$

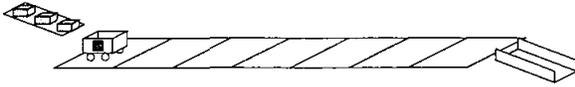

Figure 2: Load-Unload Environment

Using these formulas, the gradients of $K$ and $V$ can be calculated in $O(T)$ time. The overall gradient is

$$\nabla \hat{R}(\pi) = \frac{\sum_i \frac{(R_s(S^i) - \hat{R}(\pi))\nabla K(Y^i, Z^i, \pi) + \nabla V(Y^i, Z^i, \pi)}{\sum_j K(Y^i, Z^i, \pi^j)}}{\sum_i \frac{K(Y^i, Z^i, \pi)}{\sum_j K(Y^i, Z^i, \pi^j)}}.$$

### 3.4 Comments on the Algorithm

The estimator $R(\pi)$ essentially uses importance sampling to weigh the returns according to how likely they are under the new policy. However, the sampled returns from $X$ are replaced with the expected returns from $X$ (given the $S$ sequence) and the weighting uses knowledge of the dynamics of the known state and does not use samples to estimate those distributions. This greatly reduces the variance and results in functions $K$ and $V$ that ignore the sampled observations and actions and instead sum across all possible known states, observations, and actions.

The inner sums (over $j$) can be built up as the algorithm progresses so the entire calculation (for either the estimate or its derivative) is linear in the number of samples. We have found that conjugate gradient ascent with line search to work well for optimization.

## 4 Experiments

To demonstrate the ability of the algorithm to exploit additional domain information, we have constructed two partially-observable worlds. Each can be converted into a set of different PKMDP models, all with equivalent global dynamics, but each with different domain knowledge.

We allow for policies with memory by incorporating the memory into the state space of the process. Because the PKMDP formulation permits the dynamics of the memory to be known to the agent, this is similar to learning a finite-state controller by gradient search. Allowing policies without a fixed memory size is more difficult and not addressed in this paper.

### 4.1 Load-Unload Problem

Figure 2 shows the load-unload environment used in this paper. A cart controlled by the agent sits on a line with seven distinct positions. When the cart is in the left-most position, it receives a box if it does not already have one. When the cart is in the right-most state and is carrying a box, the box is unloaded and the agent receives one unit of reward. The agent receives zero reward in all other situations. The agent additionally has one memory bit. It may take one of the four combination of one memory action and one movement action. The memory actions are to either set or clear the memory bit while the move actions are to move the agent either to the left or to the right one position on the line. A move action which would move the agent off one end of the line leaves the agent's position unchanged. The agent can only observe the position of the cart and the value of the memory bit. It cannot observe whether the cart is loaded. The agent begins in the left-most position.

If modeled as a POMDP, this environment would have 26 hidden states, 14 observations, and 4 actions. There are a large number of different PKMDP formulations of the system. We have chosen three to illustrate differing amounts of world knowledge.

**Model 1** (No world knowledge): The variable $s$ represents the complete system state. $y$, $x$, and $o$ are all the observation (14 values), deterministically reproduced down the chain. Therefore, $p_x(x'|x, a, y)$ actually does not depend on $x$ or $a$ and is just an identity mapping (as is $p_o(o|x)$). $a$ and $z$ are the action (similarly $p_z(z|x, a)$ does not depend on $x$ and is just an identity mapping).

**Model 2** (Memory dynamics known): The variable $s$ represents the position of the agent and whether or not it is loaded. $y$ is the position of the cart and $z$ is the movement portion of the action. $x$ is the position of the cart and the memory setting. $o$ is identically the same as $x$. In this model, $p_x(x'|x, a, z)$ represents the dynamics of the memory (a simple latch). However, the position of the cart encoded in $x$ is duplicated from the variable $y$. $p_x$ does not represent the dynamics of the cart's position.

**Model 3** (End-point and memory dynamics known): The variable $s$ represents whether or not the cart is loaded, unloaded, or was just unloaded (3 values). The variable $x$ represents the position of the cart and the memory value. $y$ has only a possible value (and therefore no information). $z$ has three values: the agent is in the left-most state, the agent is in the right-most state, and the agent is in some middle state. This model encodes the knowledge that the agent is on a line and the effects of moving left or right as well as the latch dynamics of the memory bit. It also encodes the knowledge that the unknown state (*i.e.* the dynamics of loading and unloading and associated rewards) does not make a distinction among the non-edge states.

In all of these models, $r_x$ is uniformly 0 (all reward is due to $r_s$). The space of observations and actions are the same for each model and, given a mapping from observations to actions, each PKMDP has the same behavior. The only differences are in where the state transition information resides. In the first model, it all resides in $p_s$. Each successive model moves more of the information into $p_x$.

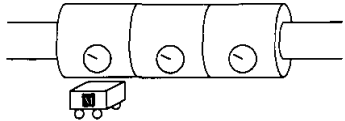

Figure 3: Clogged Pipe Environment

Our learning algorithm assumes that both $y$ and $z$ are observable (at least after the end of a trial). For these three models, that is not a problem. In every case, the values of $y$ and $z$ for a time step can be determined from the observation and action. In model 1, $y_t = o_t$ and $z_t = a_t$. In the case of model 2, $y_t$ is the position portion of $o_t$ and $z_t$ is the movement portion of $a_t$. In model 3, $y_t$ is always 1 and $z_t$ is whether the new action moves the agent to the left end, the right end, or a place in the middle. This is also completely determined (by the observation of the current position and the action taken).

### 4.2 Clogged Pipe Problem

Figure 3 shows the clogged pipe problem. A pipe, divided into three sections, is monitored by the agent. At any particular time, any subset of the three sections might be clogged. The agent is positioned next to one of the sections. It observes its position, the value of the memory bit, and whether the pipe section at its location is clogged. It may perform one of eight actions, the cross product of the spaces of memory actions and non-memory actions. The memory actions allow the clearing or setting of a single memory bit. The non-memory actions are move left, move right, wait, or unclog. The first two move the agent left or right (or leave the agent's position unchanged if such a move would take it off the end). The third action does nothing. The fourth action causes any debris blocking the current pipe section to move one section to the right (downstream) and get stuck there. The middle and right sections only become clogged if the respective "upstream" section (the section to their left) had debris and was unclogged. The left section becomes clogged based on the status of the incoming water flow. This flow has three states: clear, low debris, and high debris. At any time step it has a 0.1 chance of transitioning to an adjacent states (*e.g.* clear can change to low but not to high). In the clear state, there is no chance that new debris will enter the system. In the low state the chance is 0.3 and in the high state the chance is 0.5. The agent gets one unit of reward for every time step during which all three sections of pipe are clear of debris.

If modeled as a POMDP, this environment would have 144 states, 12 observations, and 8 actions. We have chosen three different PKMDP representations.

**Model 1** (Only memory known): Just as model 2 for the load-unload problem. The unknown state transitions encode all of the dynamics except for the memory bit.

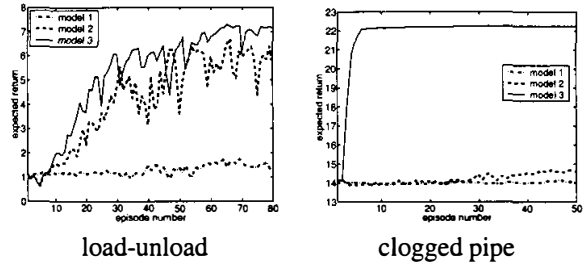

load-unload      clogged pipe

Figure 4: Results for the load-unload and clogged pipe problems.

**Model 2** (Known cart control): The agent's position and memory value are kept in the variable $x$ and their dynamics are known. The only message passed through $z$ to the unknown dynamics is which (if any) section was unclogged. The message $y$ from the unknown state indicate which sections are currently clogged. The non-zero rewards are now functions of $x$; the agent understands the goal of clearing all of the pipe sections. However, the agent does not understand the effects of clearing a pipe section in detail. It only knows that the positioning of the clearing operation is what matters to the environment.

**Model 3** (Only incoming flow unknown): The agent's position, the status of the pipes, and the memory value are all part of the known dynamics. The unknown state is only the status of the incoming flow (clear, low, high). The agent knows that its actions do not affect these values (*i.e.* $z$ can only take on one value). The message $y$ indicates whether new debris has entered into the left section. The agent still does not know the state space of the incoming flow or its effect on the upstream segment of pipe. It does know that the dynamics of the incoming flow are independent of its actions.

Once again, we have three models that encode the same dynamics, but with differing quantities of domain knowledge. Note that the second and third models allow the reward function to be moved to the known sequence. They additionally (and coincidentally) require observation of $Y$ and $Z$ which are not completely determined by the action and observation sequences.

In contrast to all previous models, models 2 and 3 require that the agent be able to gather the sequences $Y$ and $Z$ for a particular trial. In model 2, this means after an episode the agent must go back and review the clogged sensor readings for all the pipe sections at each time step. For model 3, the agent must go back and review whether new debris flowed into the left pipe for each time step. This is probably reasonable for many applications; often data is available after a trial which was not observed during the trial.

### 4.3 Results

Figure 4 shows the results of the algorithm on load-unload and clogged pipe problems. For each problem, we ran the algorithm 10 times. Each run consisted of a series of episodes (80 for the load-unload and 50 for the clogged pipe) of 100 time steps. For graphing purposes only, we took the policy used at each episode (because the algorithm is greedy, this is its current best guess at the optimal policy) and calculated its true expected return by using the whole model. This is exactly the value of the policy the algorithm would pick if it were required to stop at this episode. The two figures show this value averaged over the 10 trials as a function of episode number.

As desired, the expected return increases with added information. For both problems, there seems to be a critical piece of information which allows the problem to be solved much faster. In the load-unload problem, this is the dynamics of the memory (the difference between models 1 and 2) and for the clogged pipe problem, this is the knowledge of the pipe dynamics (the difference between models 2 and 3). However, all of the models show improved performance over less informed models.

## 5 Discussion

While some types of domain knowledge cannot be incorporated into this framework, we feel it allows for a natural and precise description of most forms of domain knowledge. In particular, it allows exact conditional probability distributions to be specified for parts of the environment without requiring that anything be known about the unknown world dynamics and state.

We find it encouraging that even the smallest bit of domain information (*e.g.* the memory bit dynamics) can make a large impact on the tractability of the problem. This gives hope that the goal of large-scale reinforcement learning system may be possible. To achieve this goal, the ideas of this paper will need to be incorporated with other new ideas for solving large reinforcement learning problems. In particular, there are better ways of doing dynamic programming that exploit structure (Guestrin et al., 2001) which would provide computational savings.

Finally, our choice of using greedy optimization for selecting the next policy is convenient and has yielded good results on a number of problems. However, this could probably be improved with a more careful trade-off between exploitation and exploration.

### Acknowledgments

This work was funded by ONR contract N00014-00-1-0637 under the MURI program "Decision Making under Uncertainty."

### References

Bengio, Y. (1999). Markovian models for sequential data. *Neural Computing Surveys*, 2, 129–162.

Dietterich, T. G. (2000). Hierarchical reinforcement learning with the MAXQ value function decomposition. *Journal of Artificial Intelligence Research*, 13, 227–303.

Guestrin, C., Koller, D., & Parr, R. (2001). Max-norm projections of factored MDPs. *IJCAI 2001* (pp. 673–680).

Hesterberg, T. (1995). Weighted average importance sampling and defensive mixture distributions. *Technometrics*, 37, 185–194.

Littman, M. (1996). *Algorithms for sequential decision making*. Doctoral dissertation, Brown University.

Maclin, R., & Shavlik, J. W. (1996). Creating advice-taking reinforcement learners. *Machine Learning*, 22, 251–281.

Ng, A. Y., Harada, D., & Russell, S. (1999). Policy invariance under reward transformations: theory and application to reward shaping. *ICML 1999* (pp. 278–287). Morgan Kaufmann.

Nikovski, D., & Nourbakhsh, I. (2000). Learning probabilistic models for decision-theoretic navigation of mobile robots. *ICML 2000* (pp. 266–274). Morgan Kaufmann.

Parr, R. E. (1998). *Heirarchical control and learning for Markov decision processes*. Doctoral dissertation, University of California at Berkeley.

Perkins, T. J., & Barto, A. G. (2001). Lyapunov-constrained action sets for reinforcement learning. *ICML 2001* (pp. 409–416). Morgan Kaufmann.

Peshkin, L., & Mukherjee, S. (2001). Bounds on sample size for policy evaluation in Markov environments. *COLT 2001*.

Peshkin, L., & Shelton, C. R. (2002). Learning from scarce experience. *ICML 2002*.

Precup, D. (2000). *Temporal abstraction in reinforcement learning*. Doctoral dissertation, University of Massachusetts, Amherst.

Precup, D., Sutton, R. S., & Dasgupta, S. (2001). Off-policy temporal-difference learning with function approximation. *ICML 2001*. Morgan Kaufmann.

Precup, D., Sutton, R. S., & Singh, S. (2000). Eligibility traces for off-policy policy evaluation. *ICML 2000*. Morgan Kaufmann.

Rabiner, L. R. (1989). A tutorial on hidden Markov models and selected applications in speech recognition. *Proceedings of the IEEE*, 77, 257–286.

Rodríguez, A., Parr, R., & Koller, D. (2000). Reinforcement learning using approximate belief states. *NIPS* (pp. 1036–1042). The MIT Press.

Rubinstein, R. Y. (1981). *Simulation and the monte carlo method*. John Wiley & Sons.

Shelton, C. R. (2001). Policy improvement for POMDPs using normalized importance sampling. *UAI 2001* (pp. 496–503).

Williams, R. J. (1992). Simple statistical gradient-following algorithms for connectionist reinforcement learning. *Machine Learning*, 8, 229–256.